# Facilitating Global Team Meetings Between Language-Based Subgroups: When and How Can Machine Translation Help?

YONGLE ZHANG, University of Maryland, USA
DENNIS ASAMOAH OWUSU, University of Maryland, USA
MARINE CARPUAT, University of Maryland, USA
GE GAO, University of Maryland, USA

Global teams frequently consist of language-based subgroups who put together complementary information to achieve common goals. Previous research outlines a two-step work communication flow in these teams. There are team meetings using a required common language (i.e., English); in preparation for those meetings, people have subgroup conversations in their native languages. Work communication at team meetings is often less effective than in subgroup conversations. In the current study, we investigate the idea of leveraging machine translation (MT) to facilitate global team meetings. We hypothesize that exchanging subgroup conversation logs before a team meeting offers contextual information that benefits teamwork at the meeting. MT can translate these logs, which enables comprehension at a low cost. To test our hypothesis, we conducted a between-subjects experiment where twenty quartets of participants performed a personnel selection task. Each quartet included two English native speakers (NS) and two non-native speakers (NNS) whose native language was Mandarin. All participants began the task with subgroup conversations in their native languages, then proceeded to team meetings in English. We manipulated the exchange of subgroup conversation logs prior to team meetings: with MT-mediated exchanges versus without. Analysis of participants' subjective experience, task performance, and depth of discussions as reflected through their conversational moves jointly indicates that team meeting quality improved when there were MT-mediated exchanges of subgroup conversation logs as opposed to no exchanges. We conclude with reflections on when and how MT could be applied to enhance global teamwork across a language barrier.

CCS Concepts: • **Human-centered computing → Human-computer interaction (HCI)**; *Empirical studies in HCI*

**Additional Key Words and Phrases:** Global teams, language choice, machine translation, shared context

**ACM Reference format:**
Yongle Zhang, Dennis Asamoah Owusu, Marine Carpuat, and Ge Gao. 2021. Facilitating Global Team Meetings Between Language-Based Subgroups: When and How Can Machine Translation Help? *Proc. ACM Hum.-Comput. Interact.* 6, CSCW1, Article 90 (April 2022), 26 pages, https://doi.org/10.1145/3512937

This work is supported by the National Science Foundation, under grant IIS-1947929.

Author's addresses: Yongle Zhang, University of Maryland, College of Information Studies (iSchool), Hornbake Library South, 4139 Campus Drive, College Park, Maryland 20742, USA, yongle@umd.edu; Dennis Asamoah Owusu, University of Maryland, Department of Computer Science, Brendan Iribe Center for Computer Science and Engineering, 8125 Paint Branch Drive, College Park, Maryland 20742, USA, dasamoah@umd.edu; Marine Carpuat, University of Maryland, Department of Computer Science, Brendan Iribe Center for Computer Science and Engineering, 8125 Paint Branch Drive, College Park, Maryland 20742, USA, marine@umd.edu; Ge Gao, University of Maryland, College of Information Studies (iSchool), 4139 Campus Drive, Hornbake Library South, College Park, Maryland 20742, USA, gegao@umd.edu.

## 1 INTRODUCTION

Global teams frequently include subgroups of people, each located in a different country and speaking a unique native language that may or may not be the team's common language (e.g., English). They serve as the backbone of modern educational programs, business projects, and research collaborations that aim to deliver worldwide impacts. Substantial evidence shows that global teams, when functioning well, will outcompete other forms of teams with regard to the scale of assessable resources (e.g., [13]), the ability to generate innovative solutions (e.g., [27]), and the performance on cognitively complex tasks (e.g., [10]). However, communication problems rooted in language diversity, or differences in native language as well as English fluency among subgroups, often prevent the team from fulfilling its potential [54].

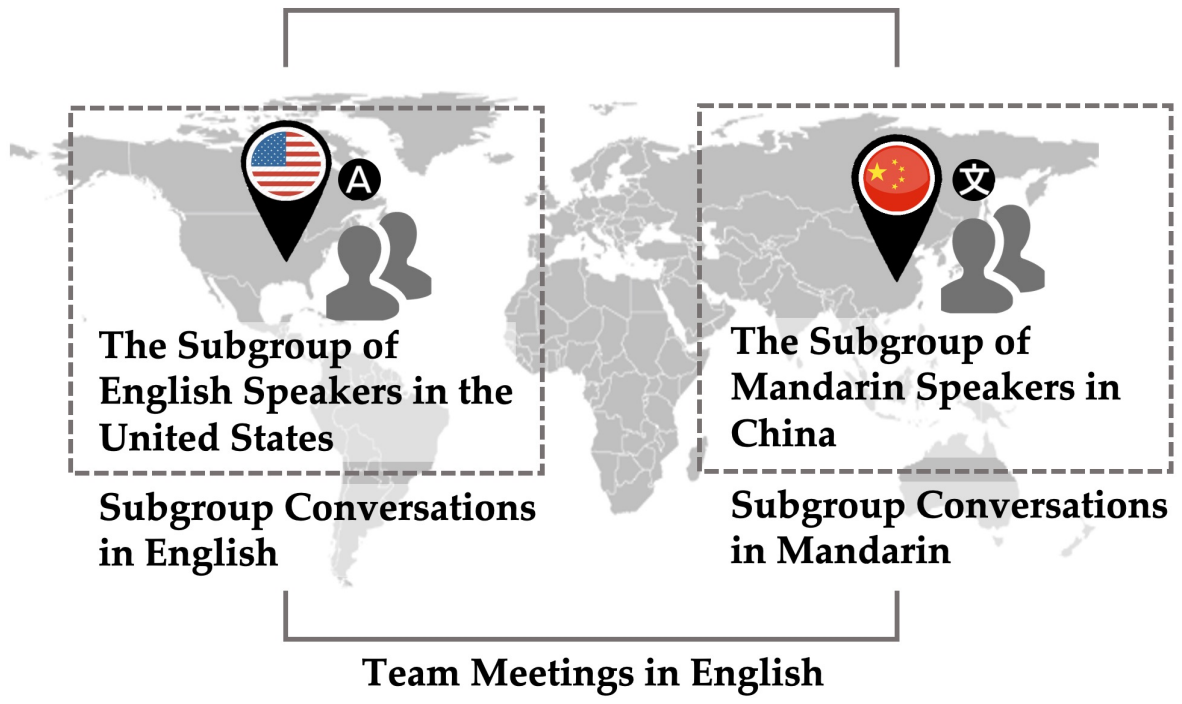

Fig. 1. A representative scenario of global teamwork between two language-based subgroups. One subgroup is located in the United States, and it involves multiple NS of English. The other subgroup lives and works in China, and it involves multiple NNS of English whose native language is Mandarin. Members within each subgroup have smooth and frequent conversations using their shared native language. However, their team meetings in English are often ineffective.

The current paper focuses on one representative scenario of global teamwork across a language barrier (e.g., [1, 32, 34, 40, 44, 50]; Figure 1). In this scenario, two subgroups of people team up to perform joint tasks. One subgroup is located in the United States, and it involves multiple native speakers (NS) of English. The other subgroup is located in China, and it involves multiple non-native speakers (NNS) of English whose native language is Mandarin. The entire team sets up periodic meetings in English to make decisions at a collective level. In preparation for those meetings, Mandarin speakers have frequent subgroup conversations using their native language, as do the English speakers. Team meetings often turn out to be challenging, especially when attendees need to assemble and act on information that was initially scattered across languages.

The above scenario reveals the complex influence of language diversity on work communication, and it has been commonly observed among global teams of various linguistic compositions (e.g., [2, 32, 34, 40, 44, 49]). A long-standing interest of CSCW and HCI scholars is to understand and facilitate global team meetings across language boundaries (e.g., [14, 18, 25, 58, 70]). However, there is a significant gap between two lines of existing research:

Studies of global teams in the field have emphasized the necessity of understanding team meetings, including challenges emerging at those meetings, as one building block of the team's entire communication flow spanning multiple languages. For instance, Cramton and colleagues found that members of each subgroup tended to perceive information communicated among themselves as taken-for-granted knowledge of the whole team. Team meetings became ineffective when subgroups each held a different mental model of what information to share and how to interpret the information shared by others [11, 12, 14]. In a more recent investigation, Hinds and colleagues conducted ethnographic research with one global team sitting across three countries. They observed that team members in each country exchanged a rich set of work-related information using the subgroup's local language. Much of this information was not accessible to speakers of a different language, which contributed to low-quality discussions and social tension at team meetings [34, 54]. These findings imply that subgroup conversations often set up the "cognitive context [46, 65]" for team meetings at a later point. Thus, pre-exchange of information from subgroup conversations is likely to benefit work communication at team meetings.

In contrast, lab studies that aimed to facilitate global teamwork usually treat team meetings as an isolated event. Recent research along this line has compared the quality of team meetings with and without real-time interventions. Examples of interventions include machine translation (MT) that converted a sentence from one language to another [67, 72], automated speech recognition (ASR) that generated multilingual transcripts for oral conversations [24], and image retrieval that indicated the semantic meaning of written messages [66]. These interventions have proved helpful in some cases but not others. One reason is probably that they only target lexical issues (e.g., whether NNS could recognize a given set of English words or sentences) that are independent of the specific context of the teamwork.

The current paper presents an experimental study that mimics real-world work communication between language-based subgroups, as illustrated in Figure 1. The contribution of this study is twofold. First, we test the hypothesis that exchanging logs of subgroup conversations before a team meeting would offer contextual information that benefits teamwork at that meeting. While this hypothesis has been implied by in-depth qualitative studies with real-world teams, there lacks supportive evidence in a quantitative format. The current study provides the missing piece. Second, we examine the idea of using MT to translate one subgroup's conversations so another subgroup can comprehend them. Underneath this idea lies a trade-off that may concern global work practitioners: MT can generate information in the team's required work language (e.g., English) at a low cost, but it can also produce outputs that are erroneous and/or incomprehensible. The current study investigates possible pros and cons of supporting team communication with MT under the given design.

To explore our hypothesis and research questions, we invited quartets of participants to perform a personnel selection task over instant messaging (IM). Each quartet consisted of two English native speakers (NS) and two non-native speakers (NNS) whose native language was Mandarin. All participants began the task with subgroup conversations in their native languages, then proceeded to team meetings using English as a common language. We manipulated the exchange of conversation logs prior to team meetings: with MT-mediated exchanges (i.e., the experimental condition) versus without exchanges (i.e., the baseline condition). We also explored whether any effects of this manipulation would vary according to participants' native language: English versus Mandarin. We measured various aspects of participants' experience at team meetings, such as their perceived quality of team communication and workload. We also collected each team's personnel selection decision and meeting logs for analytical purposes.

Consistent with our hypothesis, the data indicate improvements of participants' team meeting experience as well as task performance under the experimental condition as opposed to the baseline condition. More interestingly, participants under the experimental condition carried out discussions at greater depths, as revealed through our manual coding of team meeting logs. The majority of these results hold for both English speakers and Mandarin speakers, although the former did not rate the translations of the latter subgroups' conversation logs to be fully comprehensible. Insights gained from this study contribute to the empirical understanding of work communication between language-based subgroups. It also sheds light on the question of when and how MT can facilitate global teamwork across language boundaries.

## 2 RELATED WORK

In this section, we review four threads of prior research. We begin with research that describes language choice and the two-step flow of communication in global teams. We then examine the issue of context sharing between language-based subgroups. After that, we present existing studies and practices that use crosslingual translation to restore a shared context at the team level. Lastly, we review a broader set of research and identify the unique role of language diversity within the myriad of factors impacting global teamwork. This literature illuminates the problem of interest in our study and its potential solutions, as elaborated in later sections.

### 2.1 Language Choice and the Two-Step Flow of Communication in Global Teams

Despite the common policy of using English as a lingua franca, communication in global teams often happen in “a cocktail of languages [32].” For instance, Tange and Lauring interviewed employees at 14 Danish subunits of global companies. Interviewees reported that they chose Danish as the primary language to communicate when there were no meetings with English speaking teammates [64]. Hinds and colleagues had similar findings from observations with international teams consisting of German and English speakers. They witnessed German speakers using their native language to discuss work on a frequent basis [55]. Other studies found that language diversity not only affects how people speak at work, but it is also revealed in written communication. A survey of employees at 70 different global corporations revealed that people often generated work-related documents using the local language of each subunit. They switched to English writing only when they considered it necessary [62].

The flexibility in language choice, along with other factors, such as geographical distance, jointly shape and reinforce a unique model of work communication in global teams [1, 32,34, 40, 44, 50]. For simplicity, we refer to the model as a two-step work communication flow. One step of this flow consists of team meetings in English. These meetings serve as an essential venue for the entire team to discuss issues that cannot be resolved by any site alone, reach team-level decisions, and generate coordination plans for next steps. While emails and other formats of asynchronous communication constitute an embedded component of today’s workplace, none of them can substitute the role taken by team meetings [15, 28, 48].

In the times between team meetings, people participate in the other step of their work communication flow, that is, subgroup conversations at each site. These subgroup conversations happen much more frequently than team meetings: not only do subgroups often work on different elements of a project, but the cost of organizing global meetings can be considerably high [50, 58]. Members of the same subgroup have a strong preference to communicate in their shared native language. Such a choice often results in complex consequences. On the positive side, speaking a person’s native language allows them to think most naturally and effectively. NNS of the team’s common language (e.g., English) often leveraged their native language to articulate thoughts at the ideation stage. They identified those conversations as a strategic device to prepare themselves for team meetings in English [21]. On the negative side, the use of various languages often fragments work communication between subgroups. Members of one subgroup can be partially, or even entirely, blind to discussions within another subgroup even though the latter did not intend to conceal them [1, 34, 44].

## 2.2 Unshared Contextual Information Between Language-Based Subgroups

Tracing back to the 1990s, extensive research has documented cases where people fail to make the optimal choices regarding what and how to communicate at global team meetings [11, 12, 14]. For instance, it is common for one subgroup to disclose decisions they have made from local conversations but without explaining how they arrived at those decisions [11]. The quality of teamwork is likely to drop, especially when different subgroups need to depend on each other and/or perform cognitively complex tasks [45, 52].

Structuring subgroup conversations in different languages adds another layer of complexity to the above problem [59]. Neeley and colleagues, for example, interviewed English speakers in global project teams of various linguistic composition. Interviewees reported that they felt lost when there were non-English discussions setting up the tone for team coordination in English [54]. On a similar note, Huysman and colleagues analyzed the meeting logs of six global teams consisting of members from the United States and the Netherlands. They found that pairs of subgroups often exchanged a minimal amount of their local information at team meetings, which threatened the success of the joint work [36].

Recent literature has proposed that what people fail to establish during global teamwork is a shared cognitive context. Specifically, cognitive context refers to alternative perspectives, cause-and-effect links, definitions and scopes of decisions, and temporal updates of ideas that team members have generated for performing their joint task [46, 65]. Such information not only takes a critical role in building team-level knowhow [20]; it can also guide team members, especially those with different language backgrounds, in articulating or inferring the contextualized meaning and importance of a message [21, 37]. The two-step flow model implies that members of global teams generate and update a significant proportion of this cognitive context, if not all, during subgroup conversations. Unfortunately, they often do not have the bandwidth or awareness to detail this context at team meetings.

## 2.3 Crosslingual Translation to Restore a Shared Context

Members of monolingual teams often "store-and-forward [19, 28]" information discussed in subgroups to maintain the shared context across subgroups. For example, Malhotra and Majchrzak studied the communication practices of 55 successful distributed teams. They noticed that subgroups in those teams frequently exchanged local discussion threads among each other prior to team brainstorming meetings. People leveraged these exchanges to track and interpret the perspectives held by other subgroups. As a result, they could make the best use of team meetings and produce high-quality work outcomes [47].

For global teams, however, the store-and-forward approach turns out demanding because subgroup conversations can happen in multiple languages [2, 32, 34, 40, 44, 49]. NS employees at international corporations often complain that their colleagues at other sites "will take forever to translate [their local discussion threads or documents] to English [25]." On the flipside, NNS members on global teams reported that they felt exhausted from doing back-and-forth translations between their native language and English [21, 61].

A small but increasing number of studies have presented cases where fluent bilinguals on the global team translate conversations or documents for their colleagues. While this approach helps individuals and subgroups restore a shared context across languages, it raises other concerns. For instance, Cramton and Hinds reported that people who took the role of a bridger experienced heavy cognitive and social burdens at work [13]. Gao and Fussell further found that people often provided rudimentary translations of the source conversations because of time pressure [26].

Alternatively, global teams and organizations may recruit professional translators, but the financial cost often makes this practice unsustainable.

### 2.4 The Unique Role of Language Diversity Within Other Factors of Global Work

Language diversity is not the only factor complicating work communication in global teams. To conclude the current section, we outline previous research that investigates other aspects of team members' communication practices. Most of this research takes a cross-cultural lens, where culture is conceptualized as one or more dimensionalized beliefs and/or behavioral norms that are shared by a group of people [33, 39, 63].

For instance, Kim conducted behavioral experiments where participants thought aloud while performing problem solving tasks. Their data showed that forcing a high level of expressiveness systematically disadvantaged Asian Americans, but it did not harm the performance of European Americans [38]. This finding has inspired CSCW research that creates visualizations of an individual's communication style to share with their cross-cultural partners (e.g., [16, 42]). Gao and colleagues studied team collaboration between American and Chinese participants. Their data verified cross-cultural differences in directness of communication [29]. Communicants with a low-context style and those with a high-context style often made different interpretations of the affect conveyed by a message [22]. He and coauthors compared Canadian and Japanese participants along multiple cultural dimensions, such as informality and temporal orientation [35]. They found that raising one's awareness of cross-cultural differences could benefit global work in the case of email-based negotiations; however, not every cultural dimension matters to the same extent [31].

Language diversity poses a fundamental and, arguably, underexplored issue that is distinct from cross-cultural differences in terms of its effects on global teamwork. In the current study, specifically, we consider language diversity as a dominant factor shaping a global team's ability as well as its need to access information that is scattered across languages. We revisit the theme of cultural style in communication at the end of this paper, discussing ways to assist cross-cultural interpretation of translated information.

## 3 THE CURRENT STUDY

The current study investigates ways to facilitate global team meetings between language-based subgroups. Our literature review suggests that subgroup conversations prior to a global team meeting set up the context for communication at the meeting. However, each subgroup often generates bits and pieces of the contextual information in a different language. If there could be a method to enable crosslingual exchanges of subgroup conversations at a low cost, such exchanges would benefit team meetings.

Previous research in CSCW and HCI has implemented MT to assist real-time conversations across languages [25, 67, 72]. We conjecture that the same technique can also be applied to generate translations of one subgroup's conversation logs for another subgroup to comprehend. Notably, we choose to let MT translate subgroup conversations from non-English languages to English but not the other way around. The reason is that team-level communication usually happens in English as the required common language. Researchers have found that NNS in an English work environment often favored asymmetric translations [8, 21, 67]. They leverage translation support for message production in their native languages but prefer receiving information in English. By adopting this approach, we intended to prevent NNS from spending additional effort on translation before team meetings held in English. We also took the opportunity to explore whether asymmetric translations would matter in the same way to speakers of different native languages.

Given the above rationale, we proposed two sets of hypotheses (H) and research questions (RQ) to unpack the relationship between MT-mediated exchanges of subgroup conversation logs, an individual's native language, and work communication at team meetings:

**H1**. At team meetings, people will perceive a higher quality of team communication if there have been MT-mediated exchanges of subgroup conversation logs as opposed to no exchanges.

**RQ1**. Will the perceived quality of communication vary according to one's native language?

**H2**. At team meetings, people will achieve a better quality of teamwork performance if there have been MT-mediated exchanges of subgroup conversation logs as opposed to no exchanges.

**RQ2**. Will the quality of performance vary according to one's native language?

Further, people need to spend extra time and effort in reading other subgroups' conversations before team meetings. The added cost may or may not neutralize our proposed benefits. We asked the following questions:

**RQ3a**. Over the entire task process, will people perceive a different level of workload if there have been MT-mediated exchanges of subgroup conversation logs as opposed to no exchanges?

**RQ3b**. Will the perceived workload vary according to one's native language?

Moreover, the setting of our current study, as well as workplaces in the real-world, provides constraints on the way people interact with other subgroups' conversations. For example, people do not have unlimited time to process the information conveyed in those conversations, or they may find some of the translated sentences not fully comprehensible. Thus, there is a question of at what level people can grasp the contextual information from subgroup conversations.

We suspect a close examination of the conversational moves during team meetings may yield insights to the above question. At one extreme, people understand and process the full details of other subgroups' discussions. They are then likely to have a productive but lightweight team meeting because much of the work-related confusion has already been resolved before the meeting. At the other extreme, people only get outlines of ideas from reading other subgroups' conversations. For example, they may realize that the cognitive complexity of the task was greater than they had assumed, but they do not acquire sufficient details to unpack this complexity [59]. Such takeaways are likely to provoke team discussions at a high level of depth so that members from different subgroups can resolve the complex puzzle together. For situations between these two extremes, the characteristics of team communication may vary accordingly. With these thoughts in mind, we asked:

**RQ4a**. At team meetings, will people communicate in different ways if there have been MT-mediated exchanges of subgroup conversation logs as opposed to no exchanges?

**RQ4b**. Will the ways of communication vary according to one's native language?

## 4 METHOD

We designed a 2 × 2 between-subjects experiment to explore our hypotheses and RQs. In this experiment, we invited quartets of participants to perform teamwork in an online environment. Each team (or quartet) consisted of two NS of English and two NNS whose native language was Mandarin. The task communication happened over two successive steps: participants first had subgroup conversations in their native languages, then proceeded to team meetings using English as a common language. The task materials assigned to each language-based subgroup was written in the subgroup's native language. We made these deliberate choices so that the experiment design would reflect characteristics of real-world global teamwork and of our research interests. We manipulated the exchanges of subgroup conversations prior to team meetings: with MT-mediated exchanges (i.e., the experimental condition) versus without exchanges (i.e., the baseline condition). We also wondered whether any effects of this manipulation would vary according to participants' native language: English versus Mandarin. We collected various measures regarding participants' communication experience as well as task performance at team meetings. The rest of this section detailed each aspect of our research method.

### 4.1 Participants

We recruited 80 participants from one university in the United States. They constituted a representative sample with demographics that were comparable to participants of related research (e.g., [8, 23, 67, 70]).

Specifically, half the participants (N = 40; 20 females, 20 males) were NS of English who grew up and received the majority of their education in the United States. Their mean age was 20.57 years (SD = 2.12). They had some experience with crosslingual communication (M = 3.88, SD = 2.14 on a 7-point scale; 1 = never, 7 = very often). They were not very experienced in using translation tools or services (M = 2.95, SD = 1.34 on a 7-point scale; 1 = never, 7 = very often).

The rest of the participants (N = 40; 27 females, 13 males) were NNS of English who were currently pursuing education in the United States. Their mean duration of living in this country was 1.83 years (SD = 1.22). They used English as a second language but were not fully fluent (M = 4.68, SD = 0.97 on a 7-point scale; 1 = minimal fluency, 7 = native-level fluency). These participants all grew up in China and spoke Mandarin as their native language. Their mean age was 24.80 years (SD = 4.73). They had some experience with crosslingual communication (M = 4.13, SD = 1.02 on a 7-point scale; 1 = never, 7 = very often). They also had a moderate level of experience in using translation tools or services (M = 4.48, SD = 1.28 on a 7-point scale; 1 = never, 7 = very often).

Participants were randomly assigned to teams of unacquainted quartets. Each team consisted of two NS of English and two NNS of English whose native language was Mandarin. There were twenty teams formed in total. These teams were randomly assigned to one of the following conditions: with MT-mediated exchanges or without.

### 4.2 Task and Materials

*4.2.1 Overview.* We developed a modified version of the Personnel Selection Task (e.g., [60]). In this task, teams of participants are instructed to act as the search committee for a research lab at their university. The lab recently initiated a joint project with a partner institution in China. A research assistant (RA) position needs to be filled for this project. The search committee is tasked with recommending the most qualified job candidate for the position from a pool of four candidates. Two of these candidates are from the United States, and the other two are from China. This set up mimics a common situation in real-world global teamwork where team members with different language backgrounds access and contribute complementary resources to their joint work. In particular, the complementary resource in our task design referred to a person's ability to collect and analyze information generated in their native language.

*4.2.2 Task Information.* We designed one curriculum vitae (CV) for each of the four candidates. These CVs shared an identical structure. Specifically, each CV included seven pieces of information: the candidate's educational background (1), research experience (3), and industrial experience (3). Participants were explicitly required to weigh different pieces of information equally. For instance, a person's educational background should be considered as no more and no less important than their research experience; a piece of information about the person's research experience should be considered as no more and no less important than another piece. Thus, each candidate had between zero and seven pieces of information that made them qualified for the job.

We provided specific evaluation criteria (Table 1) that explained what a piece of qualified CV information meant for the RA position and how a candidate's total number of qualifications would be counted. Participants must have completed a training session before they could begin the formal task. The training was hosted on Qualtrics, and it required every individual participant to evaluate multiple examples of CV information. The accurate evaluation of each example was specified after participants had provided their own answers. For instance, "*Candidate X has one piece of research experience where she interviewed 3 students to ask about their experience of using an online shopping platform,*" this piece of information should not be evaluated as qualified because it does not indicate the candidate's research experience in quantitative analysis. Participants in our formal study all had provided accurate answers to the given training questions. We leveraged this design to avoid individual differences in task competency among participants.

Table 1. Criteria Guiding Participants' Evaluation of Job Candidates in the Personnel Selection Task.

| Information on the CV | Preferred Qualification | Detailed Information that Indicates the Qualification | Maximum Number |
|---|---|---|---|
| **Education Background** | Excellent coursework performance | Information under *Education Background* that indicates the candidate has received *top grades* in their education program | 1 |
| **Research Experience** | Solid experience in quantitative analysis | Information under *Research Experience* that indicates the candidate has performed *large-scale data analysis* using *statistical and/or computational methods* in research projects | 3 |
| **Industrial Experience** | Rich experience with diverse teams | Information under *Industrial Experience* that indicates the candidate has worked on *cross-departmental or multi-institutional teams* in industrial projects | 3 |
| **Sum** | - | - | 7 |

*4.2.3 Information Distribution Within Teams.* Each of the four participants on the same team (or search committee) received an exclusive subset of the candidates' CV information. We distributed a unique combination of qualified versus unqualified CV information to each participant (Table 2). This rule of design helped us achieve two purposes. First, the initial access to different but complementary information increased the necessity for participants to communicate with others and perform crosslingual exchanges. It set up a situation of interdependent and cognitively complex teamwork, where the effectiveness of a team meeting mattered [15, 28, 48, 50]. Second, the designed distribution of CV information implied an optimal choice for the job candidate (i.e., Candidate Chen). Thus, it allowed us to evaluate task performance against an objective standard. Participants could only identify wrong candidates if they considered information that was solely available to their own subgroup.

Table 2. The Distribution of Qualified Versus Unqualified CV Information Among Team Members.

| | **Candidate Yang** | | **Candidate Chen** | | **Candidate Alex** | | **Candidate Taylor** | |
|---|---|---|---|---|---|---|---|---|
| | Qualified | Unqual. | Qualified | Unqual. | Qualified | Unqual. | Qualified | Unqual. |
| **Participant 1** | 2 | 1 | 1 | 1 | 0 | 1 | 1 | 0 |
| **Participant 2** | 1 | 1 | 1 | 2 | 0 | 1 | 0 | 1 |
| **Participant 3** | 0 | 1 | 1 | 0 | 2 | 1 | 1 | 1 |
| **Participant 4** | 0 | 1 | 1 | 0 | 1 | 1 | 1 | 2 |
| **Sum** | 3 | 4 | 4 | 3 | 3 | 4 | 3 | 4 |

* *Note*: Across all the teams, participant 1 and participant 2 were always native Mandarin speakers. Participant 3 and participant 4 were always native English speakers. Among the four candidates, Yang and Chen were described as the job candidates from China. Alex and Taylor were described as job candidates from the United States. We used these pseudo names because they were gender-neutral ones.

## 4.3 Interface and System

*4.3.1 Overall Environment of the Experiment Session.* Participants performed the experiment via an IM-based platform developed by the research group. For the convenience of message translation and data collection, the platform did not afford any audio/video-based interactions among participants. Participants could reach out to the experimenter over Zoom, which permitted them to ask questions if needed. The communication between a participant and the experimenter, if happened, would be invisible to other participants. All the IM messages exchanged between participants were automatically saved on the platform's server for analysis.

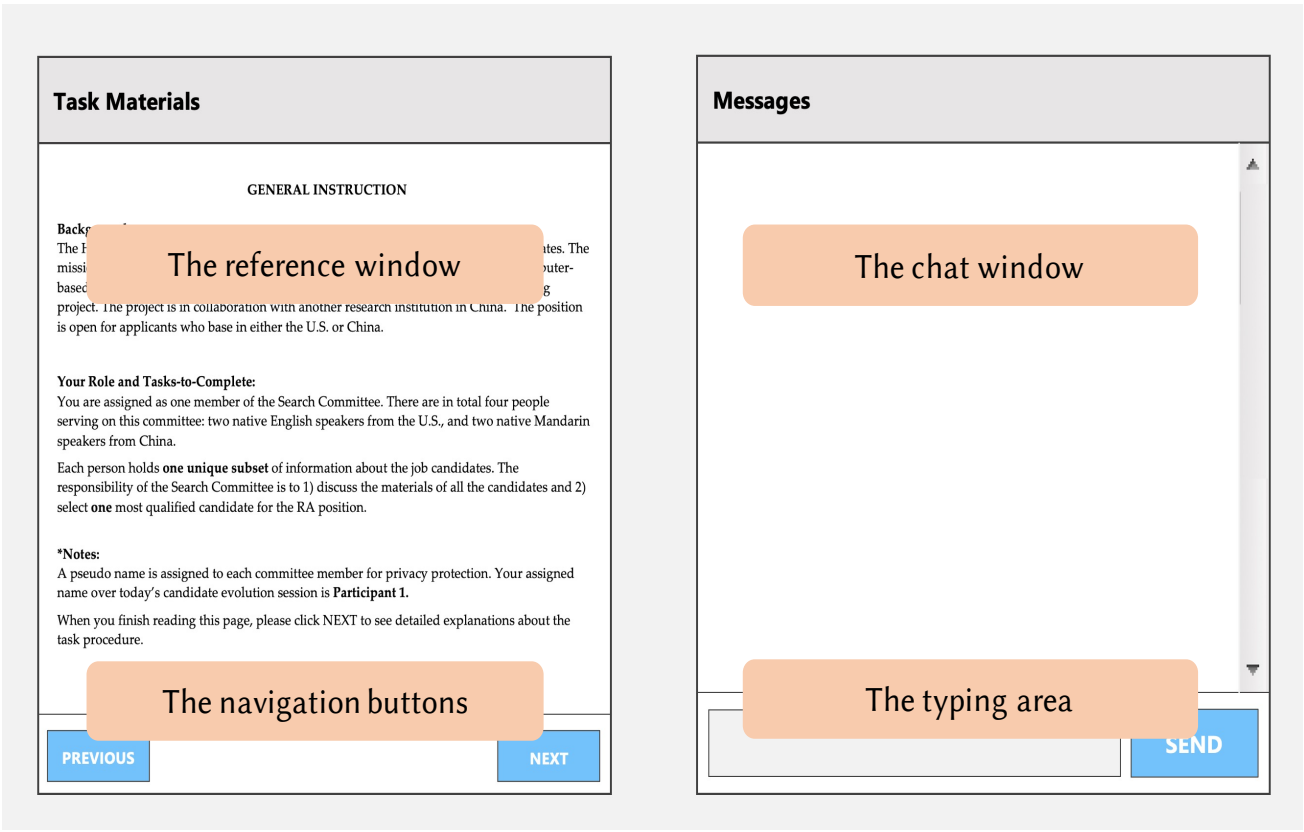

Fig. 2. The task interface consisted of a chat window and a reference window. Depending on one's task progress at each moement, the reference window displayed materials including task instructions, the candidate's CV information, links to surveys, and the other subgroup's discussion logs.

*4.3.2 IM Interface.* Participants used IM to complete subgroup conversations and team meetings. The interface followed a standard design similar to that of Slack and other messaging tools (Figure 2). Participants could see real-time messages sent by their teammates in a *chat* window. On the left side of the chat window, we placed a *reference* window. Depending on a person's current progress with the experiment procedure, the reference window displayed corresponding materials including task instructions, the candidate's CV information, links to surveys, and the other subgroup's discussion logs. With the above design, participants had convenient access to the task materials at different stages of the candidate evaluation process.

*4.3.3 MT System.* We implemented a state-of-the-art MT system using open-source tools and data to translate between Mandarin and English. The MT system adopted a neural sequence-to-sequence model that was implemented in the AWS Sockeye toolkit[1]. The model design, training configuration, and training data were based on the top performing systems in public benchmark testing. Specifically, we used an encoder-decoder model based on a 6-layer transformer network of size 512, with 8 attention heads, and a feedforward network size of 2,048. The training data comprised 17.6 million Mandarin sentences paired with their English translations, drawn from diverse news sources and United Nations corpora. The resulting system achieved a translation quality comparable with strong base transformer systems at the WMT2018 benchmark, with a BLEU score of 23.6 on the official test set. Pilot testing found that it produced natural and reasonable translations on text samples collected for the current task.

Notably, although commercial MT services (e.g., Google Translate) could also provide translation outputs for the current study, we chose to use our in-house system. This choice allowed us to improve the translation algorithm as well as its evaluations based on the current findings. The improved design enables future research as stated in later sections.

## 4.4 Procedure

Participants were assigned into unacquainted quartets consisting of two NS of English and two NNS whose native language was Mandarin. They performed the personnel selection task following a multi-step procedure as illustrated in Figure 3. From the very beginning of this task, every participant had been explicitly informed that their responsibility was to help the team identify the most qualified job candidate.

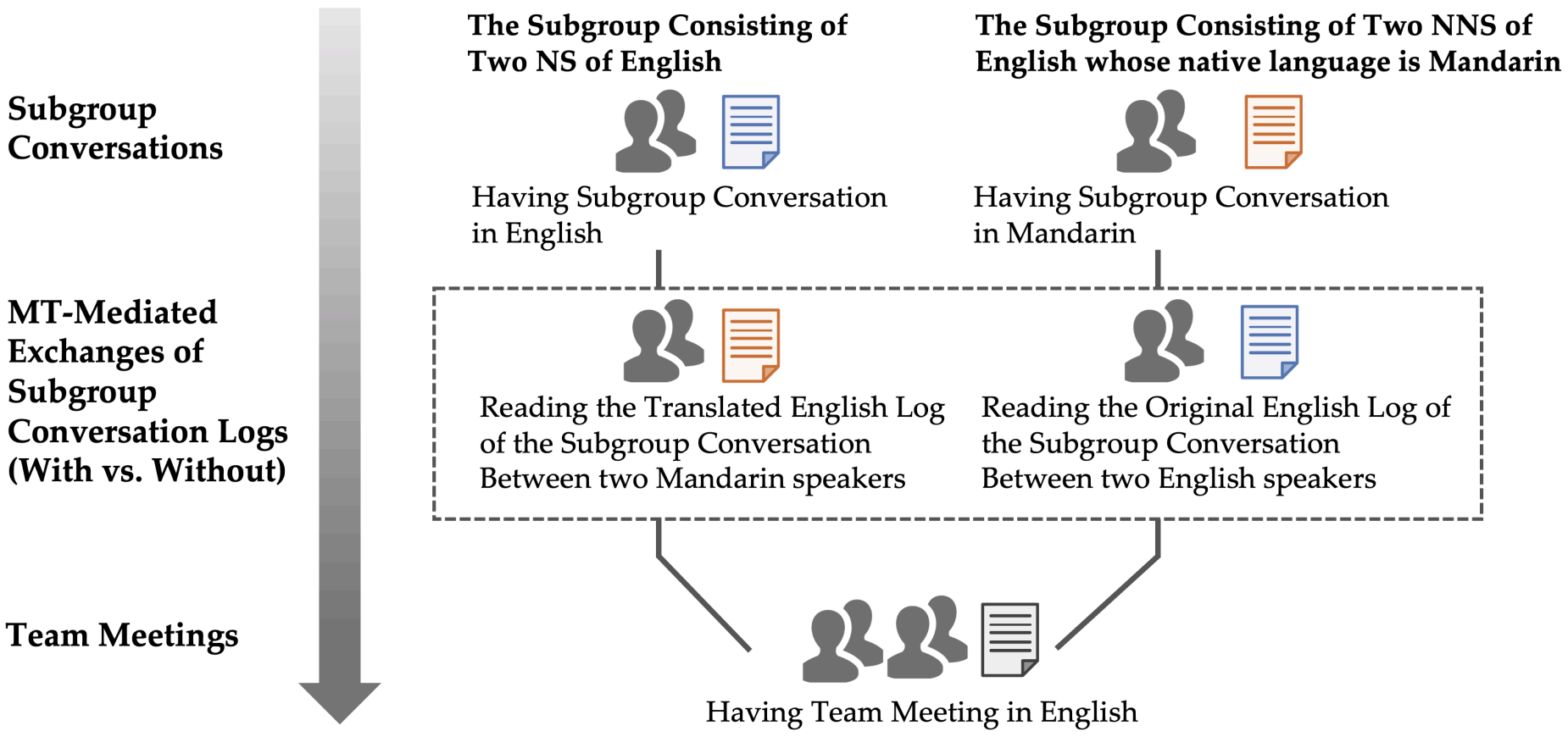

Fig. 3. Teams of participants performed the personal selection task following a multi-step procedure. All the participants began the task with subgroup conversations in their native language and they finished by attending team meetings in English. In the middle of these two steps, participants received and read the other subgroup's discussion log if their team was assigned to the experiment condition.

In the first step of the task, participants had subgroup conversations with their native speaking fellows, using the subgroup's shared native language. These conversations served as preparatory discussions for the upcoming team meetings. Each person could refer to their exclusive subset of the CV information while attending the subgroup conversations. Participants were aware that they would generate the formal candidate recommendation at a later point in time and through discussions with the entire team. Participants of the experimental condition were also aware that they would exchange their conversation log with the other subgroup prior to the team meeting. The subgroup conversation lasted for 15 minutes.

[1] https://github.com/awslabs/sockeye

The MT-mediated exchanges of subgroup conversation logs happened *only* with teams that were assigned to the experimental condition. During this step, subgroups of English speakers received an English translation of their Mandarin-speaking teammates' conversation logs. Similarly, subgroups of Mandarin speakers received the original English logs of the English-speaking teammates' conversations. These exchanges happened in nearly real-time when subgroup discussions were completed. Participants under the experimental condition were given 10 minutes to read others' conversation logs. Immediately after that, they filled out a survey hosted on Qualtrics to indicate their perceived comprehensibility of those logs. Participants under the baseline condition did not experience the exchange of subgroup conversation logs, nor did they have any surveys to fill out. They moved directly from subgroup conversations to team meetings.

In the last step of the task, participants had team meetings in quartets and using English as a required common language. The entire team discussed the information they had about the job candidates. Each person could refer to their exclusive subset of CV information, but they could not retrieve any subgroup conversations from earlier steps. All the teams were required to recommend the most qualified job candidate by the end of their meetings. The team meeting lasted for 15 minutes. Right after the team meetings, all the participants filled out a survey hosted on Qualtrics to provide feedback about various aspects of their teamwork experience.

## 4.5 Measures

We conducted three types of measures, including participants' subjective experience as indicated in their survey responses, task performance as indicated in their final recommendation of the most qualified candidate, and conversational moves at team meetings as indicated in the meeting logs. All the measures were collected at the individual level.

*4.5.1 Subjective Experience.* We measured participants' perceived *quality of communication at the team meeting* on 7-point scales (1 = strongly disagree, 7 = strongly agree). The measure was initially developed by Liu and colleagues [43]. It consisted of three multi-item subscales: *clarity* (e.g., "I understood what was said by members of the other subgroup," Cronbach's $\alpha$ = .85), *responsiveness* (e.g., "Members from the other subgroup responded to my questions and requests quickly," Cronbach's $\alpha$ = .70), and *comfort* (e.g., "I felt members of the other subgroup were trustworthy," Cronbach's $\alpha$ = .75). Participants provided their ratings on these three subscales after the team meeting. The average rating on each subscale was used for data analysis.

In the same question block of the above measure, we also asked participates to indicate *their* level of *confidence* in the team's final candidate recommendation (1 = not at all, 7 = very much). The rating was used to calibrate participants' task performance (see Section 4.5.2 for details).

We measured participants' perceived *workload of the entire task* on 7-point scales (1 = not at all, 7 = very much). The measure was adopted from the NASA TLX scale (e.g., "How much mental and perceptual activity as required, such as thinking, deciding, calculating, etc?" Cronbach's $\alpha$ = .62) [30]. Participants provided their ratings on the scales at the end of the experiment session. The average rating was used for data analysis.

We measured participants' perceived *comprehensibility of the other subgroup's discussion log* on 7-point scales (1 = strongly disagree, 7 = strongly agree), if they were assigned to the experimental condition. The measure was adopted from Liu and colleagues' clarity scale (e.g., "I understood what was said in other subgroup's discussion log," Cronbach's $\alpha$ = .83) [43].

Participants provided their ratings after reading the other subgroup's discussion log. The average rating was used for data analysis.

*4.5.2 Task Performance.* We scored each team's *final candidate recommendation* as specified in their meeting logs. Participants received a base performance score of 1 if the team selected the most qualified candidate as per our task design (i.e., Candidate Chen). Otherwise, they received a base performance score of -1. The base score was then multiplied by each participant's self-reported level of *confidence*. Thus, participants were scored between 1 and 7 if their teams successfully identified Candidate Chen as the most qualified candidate. Otherwise, they would be scored between -7 and -1. This calibration process considers the common situation where team members made their joint decision at team meetings but with different levels of certainty.

*4.5.3 Conversational Moves.* We coded all the team meeting logs to analyze how people communicated at these meetings (Table 3). The coding scheme was developed through an iterative process, following the practice recommended in previous research (e.g., [4, 7]).

Table 3. Codes Developed for Analyzing Participants' Conversational Moves at Team Meetings

| Code | Definition | Meeting Log Example | Category |
|---|---|---|---|
| **General Pointer** | Specifying the amount or type of information that is (not) provided on the candidate's CV, but without describing the detailed content | "*I only saw one (piece of information) for Taylor.*" | Content of the task communication |
| **Direct Evidence** | Describing or quoting the detailed information that is provided on the candidate's CV | "*What I have (on Alex's CV) says Alex worked closely with colleagues from the developer team, the marketing team, and the customer service team during his internship.*" | |
| **Inferred Qualification** | Describing the detailed qualification that is inferred from the candidate's CV | "*Chen has cross-departmental work experience on my side.*" | |
| **Inferred Conclusion** | Describing the overall ranking that is inferred from the candidate's CV | "*I think Yang is the best one.*" | |
| **Acknowledgement** | Indicating the acceptance of what was said in the previous turn and by the previous speaker | "*Oh, yeah.*" | Coordination of the communication process |
| **In-depth Prompt** | Guiding the previous speaker to elaborate on their previous turn | "*Do you think that experience is about big data?*" | |
| **Forward Prompt** | Guiding the team to move toward a new section of the candidate evaluation process | "*Let's move on to Yang.*" | |

First, we adopted Clark's classical framework in which he positioned all conversational moves in performing joint tasks under two broad categories: the content-related moves that attempt to carry out the official business or task, and the coordination-related moves that attempt to create a successful communication of the content (e.g., [9]). We then read through all the team meeting logs collected in the current study, identifying conversational moves that fell into each of the above categories. After multiple iterations, we arrived at a list of codes that were mutually exclusive. These codes covered the vast majority of conversational moves in the meeting logs. The only exceptions were a few moves where people corrected typos in previous turns. We excluded the typo correction moves from both the coding process and the rest of our data analysis because no additional content nor further coordination was conveyed through those moves.

We asked two fluent English-Mandarin bilinguals to code the entire dataset of team meeting logs independently. In a small number of cases, the participant issued a message that included multiple types of conversational moves. The coders would specify different codes for different parts of that turn. For example, the message "*Alex has a GPA of 3.9, and she has done statistical analysis with large-scale data*" was coded as both Direct Evidence (for the former clause) and Inferred Qualification (for the latter). The coding results produced an intercoder reliability of .84. We then discussed and resolved those inconsistent codes one by one.

## 5 RESULTS

We built a series of ANOVAs models to explore the aforementioned hypotheses and RQs. These models all reflected a 2 × 2 between-subjects design. One independent variable (IV) considered MT-mediated exchanges of the other subgroup's conversation logs: with or without; referred to as MT-mediated exchanges hereinafter. The other IV was participants' native language: English or Mandarin. Participants were nested within language-based subgroups. Subgroups were nested within teams. The Satterthwaite's approximation was applied to estimate the degrees of freedom, which often generated non-integer values.

Control variables in our analyses included participants' demographic information (e.g., gender, age) and their previous experience with crosslingual communication. We performed the heterogeneity of variance tests, comparing the variances of each control variable across groups. Results indicated that the distribution of participants' crosslingual communication experience violated the homogeneity assumption. We, therefore, substituted this variable with its log transformation: $\hat{y}_{ij} = (\log y_{ij} + 1\ )$. The transformed variable satisfied the homogeneity assumption, and it was independent of both IVs.

The rest of this section describes the main effect of each IV as well as their interactions. We also report the least squares means and standard errors for the tested variables. We do not discuss the control variables further because their effects were nonsignificant.

### 5.1 Quality of Communication at the Team Meeting

Our H1 and R1 concerned participants' perceived *quality of communication at the team meeting*. We hypothesized that people would experience a higher quality of team communication if there had been MT-mediated exchanges of subgroup conversation logs as opposed to no exchanges. Results from our data analysis supported this hypothesis.

The measurement of the quality of communication consisted of three subscales: *clarity*, *responsiveness*, and *comfort*. We conducted one separate ANOVA analysis with each aspect of this measure. The analysis indicated no significant main effect nor interaction effect on responsiveness. However, we found significant main effects of MT-mediated exchanges on clarity and comfort (Figure 4).

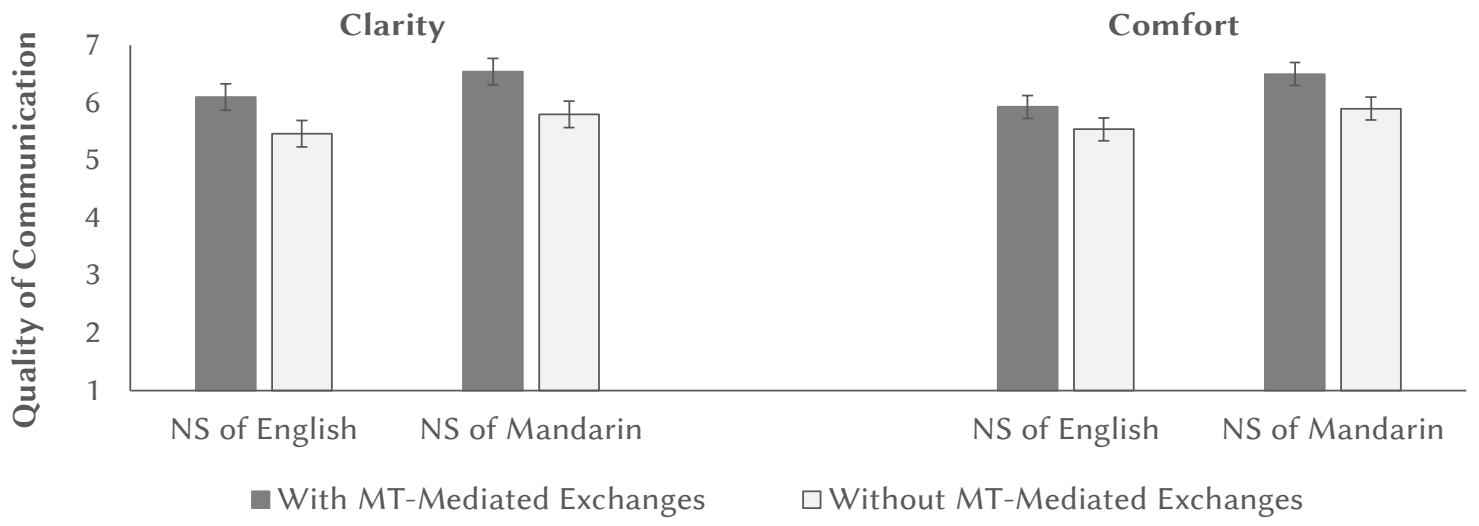

Fig. 4. Mean clarity (left) and comfort (right) by MT-mediated exchanges and native language. Together, these measures indicate that participants perceived a higher quality of team communication when there had been MT-mediated exchanges of subgroup conversation logs as opposed to no exchanges.

There was a significant main effect of MT-mediated exchanges on *clarity*: F [1, 35.43] = 12.18, $p < .01$. Participants perceived the communication with members of the other subgroup to be clearer when there were MT-mediated exchanges (M = 6.32, S.E. = .14) as opposed to no exchanges (M = 5.63, S.E. = .14). There was no significant main effect of native language on this measure: F [1, 37.51] = 2.78, $p = .10$. There was no significant interaction effect between MT-mediated exchanges and native language either: F [1, 34.84] = .06, $p = .81$.

Further, there was a significant main effect of MT-mediated exchanges on *comfort*: F [1, 35.41] = 6.37, $p < .05$. Participants perceived members of the other subgroup to be more trustworthy when there were MT-mediated exchanges (M = 6.22, S.E. = .14) as opposed to no exchanges (M = 5.72, S.E. = .14). The main effect of native language on this measure was marginally significant: F [1, 38.36] = 4.06, $p = .05$. NS of Mandarin provided higher ratings (M = 6.20, S.E. = .15) than NS of English (M = 5.73, S.E. = .15). However, there was no significant interaction effect between MT-mediated exchanges and native language: F [1, 34.85] = .26, $p = .61$.

In sum, the above results suggested that our manipulation of the MT-mediated exchanges enhanced people's perceived *clarity* and *comfort* at team meetings regardless of native language.

### 5.2 Task Performance on the Personnel Selection Task

Our H2 and RQ2 concerned participants' *task performance at the team meeting*. We hypothesized that people would reach a higher quality of teamwork performance if there had been MT-mediated exchanges of subgroup conversation logs as opposed to no exchanges. Results from our data analysis partially supported this hypothesis.

We first compared the *final candidate recommendation* given by each team. In the baseline condition without exchanges of subgroup discussion logs, none of the 10 teams identified Candidate Chen as the most qualified one. In the experimental condition with MT-mediated exchanges, 4 out of the 10 teams identified Candidate Chen as their final recommendation.

We then conducted an ANOVA analysis with the *calibrated task performance score* of each participant. There was a significant main effect of the MT-mediated exchanges on *this measure*: F [1, 36.04] = 16.64, $p < .01$. Participants performed the task more successfully when there were MT-mediated exchanges (M = .28, S.E. = 1.06) as opposed to no exchanges (M = -5.81, S.E. = 1.06). The was no significant main effect of native language on this measure: F [1, 38.97] = .01, $p = .94$.

There was no significant interaction effect between MT-mediated exchanges and native language either: F [1, 36.22] = .01, $p$ = .93.

The mean calibrated task performance scores imply that participants might feel highly confident no matter whether their team had identified the objectively best candidate or not. To verify this conjecture, we performed two additional ANOVA analyses with participant's self-reported *confidence* along, first across all participants, then of those within the experimental condition.

When comparing participant's self-reported *confidence* across all participants, those who performed the task with MT-mediated exchanges (M = 6.22, S.E. = .17) reported a slightly higher level of confidence than those with no exchanges (M = 5.85, S.E. = .17), but the difference was not significant: F [1, 35.26] = 2.32, $p$ = .14. NS of English (M = 6.08, S.E. = .18) and NS of Mandarin (M = 6.00, S.E. = .18) did not show significant difference on this measure: F [1, 38.76] = .09, $p$ = .77. There was no significant interaction effect between MT-mediated exchanges and native language either: F [1, 34.74] = .02, $p$ = .90.

We then compared participant's self-reported *confidence* within the experimental condition where MT-mediated exchanges happened. Participants whose team selected the objectively best candidate (M = 6.57, S.E. = .25) reported a slightly higher level of confidence than those in other teams (M = 5.88, S.E. = .25), but the difference was not significant: F [1, 14.63] = 3.76, $p$ = .07. NS of English (M = 6.28, S.E. = .27) and NS of Mandarin (M = 6.17, S.E. = .27) did not show significant difference on this measure: F [1, 19.17] = .08, $p$ = .78. There was no significant interaction effect between candidate selection and native language either: F [1, 16.35] = .65, $p$ = .43.

Thus, the results suggested that our manipulation of the MT-mediated exchanges enhanced *task performanc*e at the team level. However, participants in general held a high-level of *confidence* in their team's final candidate recommendation. They tended to believe that their team had identified the objectively best candidate even when it was not true.

## 5.3 Workload of the Entire Task

Our RQ3a and RQ3b considered participants' perceived *workload of the entire task*. The ANOVA analysis indicated no significant main effect nor interaction effect.

Specifically, the workload did not appear to vary when there were MT-mediated exchanges (M = 3.52, S.E. = .15) as opposed to no exchanges (M = 3.58, S. E. = .15): F [1, 35.21] = .10, $p$ = .75. NS of English (M = 3.50, S.E. = .16) and NS of Mandarin (M = 3.60, S.E. = .16) did not perceive their workload to be significantly different: F [1, 38.73] = .17, $p$ = .68. The interaction effect between MT-mediated exchanges and native language was not significant: F [1, 34.70] = 1.77, $p$ = .20.

## 5.4 Ways of Communication at the Team Meeting

Our RQ4a and RQ4b targeted participants' *ways of communication at the team meeting*. The coding of the team meeting logs yielded seven types of conversational moves: *general pointer*, *direct evidence*, *inferred qualification*, *inferred conclusion*, *acknowledgement*, *in-depth prompt*, and *forward prompt*. We conducted a separate ANOVA analysis with each type of these moves.

The analyses indicated no significant main effect nor interaction effect for use of direct evidence, inferred conclusion, acknowledgement, and forward prompt. However, we found significant main effects of MT-mediated exchanges on general pointer, inferred qualification, in-depth prompt, as well as reasoning pair, or the co-existence of direct evidence and inferred qualification (Figure 5).

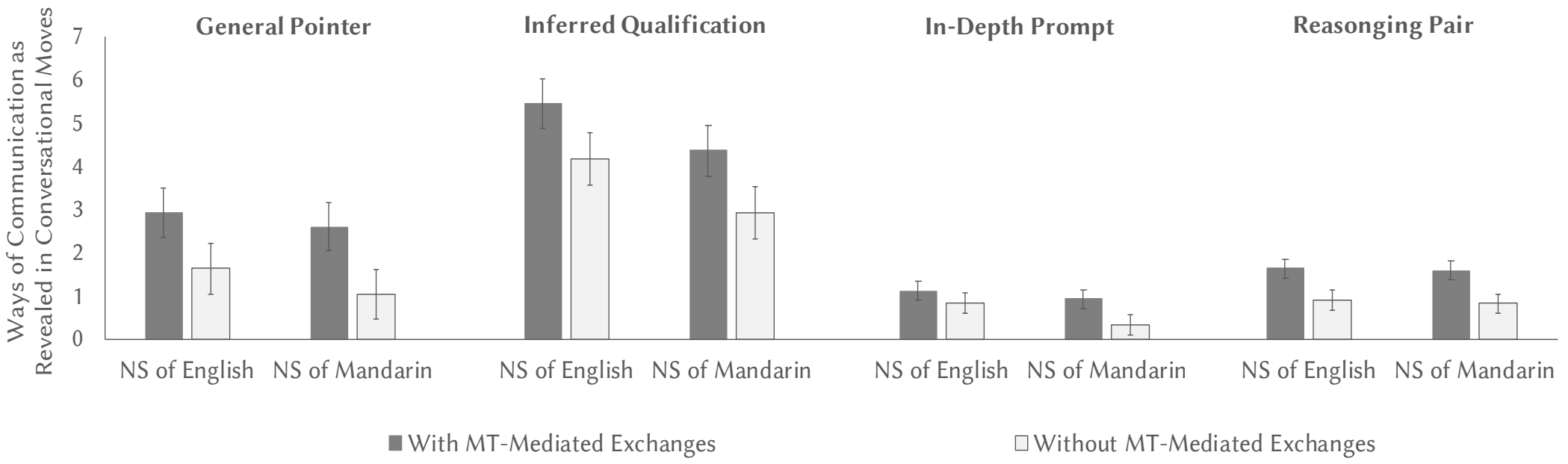

Fig. 5. Mean use of general pointer, inferred qualification, in-depth prompt, and reasoning pair, respectively, by MT-mediated exchanges and native language. Together, these conversational moves indicate that participants structured team discussions of a greater level of depth when there had been MT-mediated exchanges of subgroup conversation logs as opposed to no exchanges.

There was a significant main effect of MT-mediated exchanges on *general pointer*: F [1, 34.25] = 6.74, $p$ = .01. Participants issued a larger number of general pointers when there were MT-mediated exchanges (M = 2.76, S.E. = .38) as opposed to no exchanges (M = 1.34, S.E. = .38). There was no significant main effect of native language on this measure: F [1, 38.91] = .53, $p$ = .47. There was no significant interaction effect between MT-mediated exchanges and native language either: F [1, 35.41] = .07, $p$ = .80.

There was a significant main effect of MT-mediated exchanges on *inferred qualification*: F [1, 33.41] = 5.58, $p$ < .05. Participants issued a larger number of inferred qualifications when there were MT-mediated exchanges (M = 4.90, S.E. = .40) as opposed to no exchanges (M = 3.55, SD = .40). There was no significant main effect of native language on this measure: F [1, 36.39] = 3.04, $p$ = .09. There was no significant interaction effect between MT-mediated exchanges and language background either: F [1, 32.85] = .02, $p$ = .89.

There was a marginally significant main effect of MT-mediated exchanges on *in-depth prompt*: F [1, 34.35] = 4.04, $p$ = .05. Participants issued a larger number of in-depth prompts when there were MT-mediated exchanges (M = 1.02, S.E. = .15) as opposed to no exchanges (M = .58, S.E. = .15). There was no significant main effect of native language on this measure: F [1, 37.97] = 1.68, $p$ = .20. There was no significant interaction effect between MT-mediated exchanges native language either: F [1, 34.35] = .51, $p$ = .48.

In addition, we examined the use of *reasoning pair* based on the developed codes. This measure referred to the co-existence of *direct evidence* and its corresponding *inferred qualification* within two adjacent turns issued by the same participant, or, in a few cases, within the same turn. Our analysis indicated a significant main effect of MT-mediated exchanges on *reasoning pair*: F [1, 34.96] = 12.25, $p$ < .01. Participants issued a larger number of reasoning pairs when there were MT-mediated exchanges (M = 1.61, S.E. = .15) as opposed to no exchanges (M = .87, S.E. = .15). There was no significant main effect of native language on this measure: F [1, 38.97] = .06, $p$ = .81. There was no significant interaction effect between MT-mediated exchanges and native language, either: F [1, 33.95] = .00, $p$ = .97.

Overall, the above results suggested that our manipulation of the MT-mediated exchanges triggered adjustments in people's *ways of* communication *at team meetings* regardless of native language. These adjustments were demonstrated through participants' increased use of *general pointer*, *inferred qualification*, *in-depth prompt*, and *reasoning pair* during team communication.

### 5.5 Comprehensibility of the Other Subgroup's Discussion Log

For participants of the experimental condition, we analyzed their perceived *comprehensibility of the other subgroup's conversation logs*. While this analysis was not in response to any particular hypotheses or RQs, it offered some clue about our MT system's performance in the current study.

Our analysis indicated a significant main effect of native language on comprehensibility: $F [1, 35] = 11.85$, $p < .01$. NS of Mandarin's comprehensibility ratings of the original English logs (M = 5.93, S.E. = .28) were higher than NS of English's ratings of the translated English logs (M = 4.46, S.E. = .28). That is, Mandarin speakers encountered few problems comprehending other subgroups' English conversations, whereas English speakers found the translations of other subgroups' Mandarin conversations to be comprehensible but not to the full extent.

## 6 DISCUSSION

To recap, the overarching goal of the current study is to facilitate global team meetings between language-based subgroups. We designed the intervention where an MT system translated logs of subgroup conversations into the team's common language (i.e., English), if needed. We found that the quality of team meetings in English improved when there had been MT-mediated exchanges of subgroup conversation logs (i.e., the experimental condition) as opposed to no exchange (i.e., the baseline condition). The improvement was jointly reflected through multiple measures, and it held for both English speakers and Mandarin speakers. In the rest of this section, we discuss three aspects of our research findings: characteristics of high-quality team meetings, benefits of information exchange between subgroups, and implications for utilizing MT.

### 6.1 Characteristics of High-Quality Team Meetings

At team meetings using English, participants in the experimental condition achieved improved task performance and enhanced communication experiences compared with their counterparts in the baseline condition. Further analyses of the meeting logs revealed several characteristics of team communication that differentiated higher-quality meetings from lower-quality ones.

One such characteristic was an increased use of general pointer, that is, conversational moves outlining what was (not) contained in a team member's exclusive subset of the CV information. Unlike other types of content-related moves, general pointer did not directly contribute to the team's evaluation of candidates. The primary function of these moves, instead, was to set up the team's shared meta-cognition of who knew what. Team members with shared meta-cognition are more likely to form reasonable anticipations of each other's needs and actions, thereby achieving their joint task more effectively [52].

A second characteristic was an increased use of inferred qualification, that is, conversational moves detailing a person's interpretation of the CV information. These moves may either benefit or hinder the joint teamwork, depending on how they are used. In some cases, for example, a piece of raw information on the candidate's CV has already been disclosed at an earlier moment in the team meeting. It is then helpful to have someone specify the qualification inferred from that information. These specified inferences align team members toward a shared task logic or remind them of misalignments that they would otherwise fail to detect. In other cases, team members may share the inferences they draw with each other, without having explicitly grounded those interpretations on the task materials. Low-quality discussions are then likely to happen because people are falsely positive about consensus established at the team level.

Table 4. Two Meeting Excerpts from the Experimental Condition (Left) and the Baseline Condition (Right).

| | **Experimental Condition** | **Baseline Condition** |
|---|---|---|
| **Turn 1** | Speaker 1: *"Taylor has one qualified research experience. 2 million figure data analysis."* | Speaker 1: "*How about Alex?*" |
| **Turn 2** | Speaker 2: *"He also has experience on a product team and designed prototypes."* | Speaker 2: "*Alex has a GPA of 3.9, and she has done statistical analysis with large-scale data.*" |
| **Turn 3** | Speaker 3: *"But did it mention working with other teams/departments?"* | Speaker 3: "*His industrial experience is cross department.*" |
| **Turn 4** | Speaker 2: *"No."* | Speaker 4: "*Ah okay.*" |
| **Turn 5** | Speaker 3: *"Then probably not (qualified)."* | Speaker 1: "*In that case, I would think Alex is one of the best.*" |
| **Turn 6** | Speaker 2: "Oh yeah." | Speaker 3: "*Yeah.*" |

We believe participants in our experimental condition used inferred qualification in a more constructive way than their counterparts in the baseline condition. This claim is supported by another two characteristics observed from the team meeting logs under the experimental condition: an increased use of in-depth prompt, or conversational moves guiding others to elaborate on their previous turn; and an increased use of reasoning pair, or the joint moves of direct evidence and its corresponding inferred qualification issued by the same person.

Table 4 offers a pair of examples that demonstrate how participants issued these conversational moves in the experimental condition and the baseline condition, respectively. Participants in the experimental condition presented their inferred qualifications by contextualizing them in reasoning pairs (e.g., Experimental Condition, Turn 1). They also leveraged in-depth prompts to dive deeper into the qualifications inferred by others (e.g., Experimental Condition, Turn 3). These conversational moves bring team members on the same page about the cognitive process required for candidate evaluation. As a result, people perceive team communication to be clear and comfortable despite their differences in native language and initial access to the task materials. In contrast, participants in the baseline condition often left the relationship between a piece of raw information and its corresponding qualification unspecified (e.g., Baseline Condition, Turn 2 – 4). Under these circumstances, discussions of candidate evaluation move forward smoothly but perhaps without being grounded on a solid base.

### 6.2 Benefits of Information Exchanges Between Subgroups

Previous research has showed that global team members often fail to "bridge different thought worlds [17]" at team meetings. The data collected in the current study echoes this point. We see that all teams shared information and drew conclusions at English team meetings. However, they may not establish common ground at the cognitive level. The MT-mediated exchanges can contribute to grounding at the team level through two mechanisms.

First, conversation logs from the other subgroup may elicit people's awareness of possibly unshared perspectives, task logics, or reasoning styles prior to team meetings. This increased awareness triggers team members to disclose a larger amount of seemly redundant information (e.g., more general pointers) at the team meeting. Prior research has reported a similar strategy in that domain experts offer over-explanations to ensure the quality of communication with laypeople (e.g., [6]). In both scenarios, participants suspect they may not share the same mental model regarding the target task or conversation topic with other communicants. They take additional steps to prevent miscommunications.

Second, conversation logs from the other subgroup may help people pinpoint the cognitive complexity of their joint task prior to team meetings. To cope with this complexity, team members carry out collective discussions in an elaborative way (e.g., more in-depth prompts). Our study yields no significant difference in people's perceived workload between conditions. Therefore, we infer that the effort spent on reading exchanged conversation logs and scrutinizing candidates' qualifications pays off, as people feel they can better unpack the complex puzzle as a team.

Further, our data imply that, although participants in the experimental condition were more likely to enter team meetings with a shared cognitive context across subgroups, it was only at a cursory level. Participants might be on the same page about potential gaps between perspectives or task logics held by different team members, as well as that the task would be too complex to resolve without in-depth discussions. However, they had not processed other subgroups' conversations in detail. Such a possibility is evident from the equal amount of direct evidence as well as inferred conclusions issued across conditions. While the pre-exchange of subgroup conversation logs can offer contextual information that benefits team meetings, it cannot substitute team meetings themselves.

### 6.3 Implications for Utilizing MT

With the above understanding of global team meetings and their relationship with subgroup conversations, we reflect on its implications for utilizing MT. Previous CSCW and HCI research has leveraged MT to enable multilingual conversations where all the interlocutors produce messages in their native languages (e.g., [8, 25, 67, 72]). Our current study builds upon that research but offers complementary insights.

Specifically, we leverage MT to offer asynchronous support for global teams across a language barrier, which is supported by empirical evidence indicating a two-step communication flow in those teams. We propose that an adequate way to use MT is to translate subgroup conversations scattered among various languages. We also verify that the pre-exchange of subgroup conversation logs can benefit task communication at team meetings. This proposal differs from most previous applications of MT whereby researchers use MT to translate ongoing conversations in a synchronous manner.

The asynchronous use of MT provides unique opportunities to improve the translation quality for the task at hand. For instance, an MT system could condition its translation of a given message upon other messages in the same conversation logs, which improves discourse cohesion [3]. The MT system could further be provided with tailored translations of key terms, which encourages precise and consistent lexical selections throughout the conversation logs [41]. When the monolingual conversation logs of a subgroup are available, it is also possible to build adaptive MT using those logs [69]. However, applying any of the above techniques to MT in real-time conversations is challenging, as model adaptation and broader context processing are both computationally taxing. The translation would be slowed down significantly.

Moreover, the current study demonstrates MT's potential to restore a team's shared cognitive context. English speakers in our experimental condition did not rate the translations of Mandarin conversation logs to be fully comprehensible. Still, those logs fulfilled their job in preparing participants for the upcoming team meetings. Based on these findings, we suspect that the context-focused approach puts a less strict requirement of the translation outputs' comprehensibility. In contrast, traditional ways of using MT often requires high precision translations because the promise is to have MT address lexical issues. Even occasional translation errors could disrupt team communication in the latter case [8, 23, 71].

Lastly, we adopted an asymmetric design of MT that translates non-English conversation logs into English but not the other way around. While recent CSCW research has implied the value of such a design to NNS of English, no empirical testing was explicitly done [8, 21, 67]. In our sample, NNS participants were Mandarin speakers who spoke English as their second language and, on average, at a medium level of fluency. We found that the pre-exchange of subgroup conversation logs in (translated) English enhanced the team meeting experience of both English speakers and Mandarin speakers. This finding supports the proposal of using the asymmetric translation design to assist team communication in an English environment.

## 7 LIMITATIONS AND FUTURE DIRECTIONS

Findings presented in the current paper are qualified by our methodological choices. In particular, we conducted experiments where participants were assigned to teams and task conditions. This method enables us to investigate phenomena of our empirical interests effectively [5]. However, it does not simulate the full dynamics of teamwork in the real world.

We reflect on four specific choices in the study design: the communication medium, the availability of subgroup conversation logs, the team composition and the task setting, and the underexplored influence of cultural style on MT-mediated communication. We elaborate on how each choice may affect the ecological validity of our findings. We also outline relevant directions where future research should explore.

### 7.1 Communication Medium

Participants in the current study had both subgroup conversations and team meetings over IM. While this choice echoes CSCW literature indicating the wide adoption of IM for work communication (e.g., [53, 73]), it leaves out other popular mediums at real-world workplaces.

Previous studies with global teams have pointed out the pros and cons of using IM as the exclusive medium to run experiments [22, 67, 72]. On the positive side, IM generates convenient, reviewable, and comprehensive logs of the communication between participants. Researchers, therefore, can draw quantitative insights by analyzing conversation logs at the message level. These insights complement qualitative findings provided by field studies, where logging all work conversations verbatim is often not possible. On the negative side, IM is considered as one representative form of lean mediums. It, therefore, limits the generalizability of our reported findings in situations where a rich medium, such as audio conferencing, is used.

Future studies should examine how MT-mediated exchanges can be enabled in work communication using non-IM mediums. For instance, researchers may apply speech recognition technology to transcribe team members' utterances for translations (e.g., [25]). They may also motivate team members to manually correct transcription errors, if any, (e.g., [24]) so that the translation outputs are more comprehensible. The common lags between subgroup conversations and team meetings in real-world settings give people the time to perform manual corrections.

### 7.2 Availability of Subgroup Conversation Logs

The current study introduces an experiment setting where all of the subgroup conversations are available for translations and/or exchanges. However, it would be inappropriate to claim the take-for-granted availability of conversation logs in real-world settings.

Previous research in HCI and CSCW has outlined a few reasons that may prompt people to conceal their subgroup conversations from the rest of the team. For example, Gao and coauthors found that NNS in an English environment often used their native languages to discuss personal matters as well as work-related issues that did not immediately connect to NS colleagues' tasks at hand. Most of these conversations were made private by lowering the communicants' voices or using non-public channels [21, 26]. Besides, members of the same subgroup may initiate informal work conversations in a spontaneous manner (e.g., watercooler chats), which results in unintended challenges for information sharing at team level. The situation can be furtherly complicated if the local policies regulating each subgroup's information disclosure are different (e.g., [13]).

We call for future studies that explore a selective translation and/or information exchange between language-based subgroups. The rationale is that people may not want to share all the local information out of their own subgroup, as well as that they may be overloaded if too much information is shared for them to process. The setting of our current study, along with our findings, indicates the importance of prioritizing information that is task-orientated, contributes to a team's shared meta-cognition (e.g., general pointers), and reveals the rationale of one's argument (e.g., reasoning pairs). Future research may investigate computational methods to auto detect the above types of information and recommend people to share them at the team level. As a side benefit of this selective approach, the time people spend on reading other subgroups' conversation logs would go down.

### 7.3 Team Composition and Task Setting

The participants in our study featured specific demographics. For example, all the teams consisted of English speakers and Mandarin speakers; there were no participants of other native language backgrounds. Mandarin speakers, on average, self-reported a medium level of English fluency. While the demographics of our sample are comparable to those in other experiment studies (e.g., [8, 23, 67, 70]), the linguistic composition of real-world global teams is likely to be more complex.

Similarly, the task setting in our study featured a few unique characteristics. Our given scenario of personnel selection required participants to pursue the team's goal collaboratively. Interpersonal dynamics under this task setting are usually different from those under other settings, such as conflict management or negotiations [51]. Our manipulation required teams under the experimental condition to spend extra time reading subgroup conversation logs, which might create potential influence on participants' exposure to the job candidates' information. Also, the best candidate in our task design had qualified experiences only marginally higher than the other candidates. Thus, the difficulty level of this task may be high for participants in general, which echoes the results we saw regarding task performance. Future studies should test the effect of MT-mediated exchanges among team composition and task settings of other variations.

### 7.4 Cultural Style in MT-Mediated Communication

Last but not least, we revisit the theme of cultural style in communication (referred to as cultural style hereinafter). We conceptualize the differences in cultural style and native language as two issues that affect global teamwork in separate ways. Previous research (see Section 2.4) demonstrates how communicants featuring contrasting cultural styles may fail to ground their interpretations of the same piece of information. It usually considers situations where the information has already been accessible at the team level. Language diversity, however, can disrupt the accessibility to information in the first place. The current study offers a proof-of-concept for our central claim revolving around the issue of language diversity, that is, the MT-mediated exchanges of subgroup conversation logs prior to a team meeting can offer contextual information that benefits teamwork at the meeting. Future studies should deepen this line of investigation, exploring how cross-cultural differences would matter to an individual's message interpretation after the message has been translated for exchanges.

Notably, investigating the above questions will not be easy for a few reasons. For instance, the differences in people's cultural style can be described from multiple dimensions (e.g., [33, 35, 56]). To date, there lacks a conclusive understanding of which dimensions would matter most to each specific task setting and/or team composition. Further, reflecting (or adjusting) a source message's cultural style in its cross-lingual translation is an ambitious goal for MT. Some recent work in this space has taken initial steps to tune the level of formality of translation outputs through computational methods [57, 68]. However, the sociotechnical gap between what real-world teams need and what current MT systems can afford remains significant. It points to an exciting space where CSCW scholars and technical experts on MT should work together. Co-design activities involving different stakeholders of global teamwork could be a good starting point to embark on this long-term and collaborative research endeavor.

## 8 CONCLUSION

Global teams frequently feature language diversity, or differences in native language as well as English fluency among subgroups. Team meetings can be challenging when attendees need to assemble and act on information that was initially scattered across languages. In this paper, we presented findings from a between-subjects experiment that mimics real-world global teamwork between language-based subgroups. We invited twenty quartets of participants to perform a personnel selection task in an online environment. Each team (or quartet) included two English native speakers (NS) and two non-native speakers (NNS) whose native language was Mandarin. Participants began the task with subgroup conversations in their native languages, then proceeded to team meetings using English as a common language. We manipulated the exchange of conversation logs prior to team meetings: with MT-mediated exchanges (i.e., the experimental condition) versus without exchanges (i.e., the baseline condition). We also compared possible effects of this manipulation, if any, on participants speaking different native languages: English versus Mandarin. We found that team meeting quality improved when there were MT-mediated exchanges of subgroup conversation logs as opposed to no exchanges. This improvement was evident from the analysis of participants' subjective experience, task performance, and conversational moves at team meetings. Our findings contribute to the empirical understanding of work communication in global teams. In particular, they highlight the crucial but often ignored role of selective information, such as general pointers and reasoning pairs, in constructing a shared cognitive context across language-based subgroups of the same global team. They also indicate the promise of using MT to facilitate global teamwork through asynchronous, context-focused, and asymmetric ways.

## ACKNOWLEDGMENTS

This work is supported by National Science Foundation, under grant IIS-1947929. We thank Shaan Chopra and Emily Gong for their assistance. We also thank the anonymous reviewers for their valuable comments on earlier versions of this paper.